\documentclass[conference]{IEEEtran}
\IEEEoverridecommandlockouts
\usepackage{cite}
\usepackage{amsmath,amssymb,amsfonts}
\usepackage{algorithmic}
\usepackage{graphicx}
\usepackage{textcomp}
\usepackage{xcolor}
\def\BibTeX{{\rm B\kern-.05em{\sc i\kern-.025em b}\kern-.08em
    T\kern-.1667em\lower.7ex\hbox{E}\kern-.125emX}}
\begin{document}

\title{Detecting Production Phases Based on Sensor Values using 1D-CNNs
}

\author{\IEEEauthorblockN{
Burkhard Hoppenstedt$^{1,}$,
Manfred Reichert$^{1}$,
Ghada El-Khawaga$^{1}$,
Klaus Kammerer$^{1}$,\\
Karl-Michael Winter$^{2}$,
 and
R\"udiger Pryss$^{3}$}
\IEEEauthorblockA{$^{1}$Institute of Databases and Information Systems, Ulm University, Ulm, Germany}
\IEEEauthorblockA{$^{2}$Research and Development, Nitrex Metal Inc., St. Laurent, Canada}
\IEEEauthorblockA{$^{3}$Institute of Clinical Epidemiology and Biometry, University of W\"urzburg, W\"urzburg, Germany}
}

\maketitle

\begin{abstract}
In the context of Industry 4.0, the knowledge extraction from sensor information plays an important role. Often, information gathered from sensor values reveals meaningful insights for production levels, such as anomalies or machine states. In our use case, we identify production phases through the inspection of sensor values with the help of convolutional neural networks. The data set stems from a tempering furnace used for metal heat treating. Our supervised learning approach unveils a promising accuracy for the chosen neural network that was used for the detection of production phases. We consider solutions like shown in this work as salient pillars in the field of predictive maintenance.
\end{abstract}


\section{Introduction}
\begin{figure}[h]
\includegraphics[width=\linewidth]{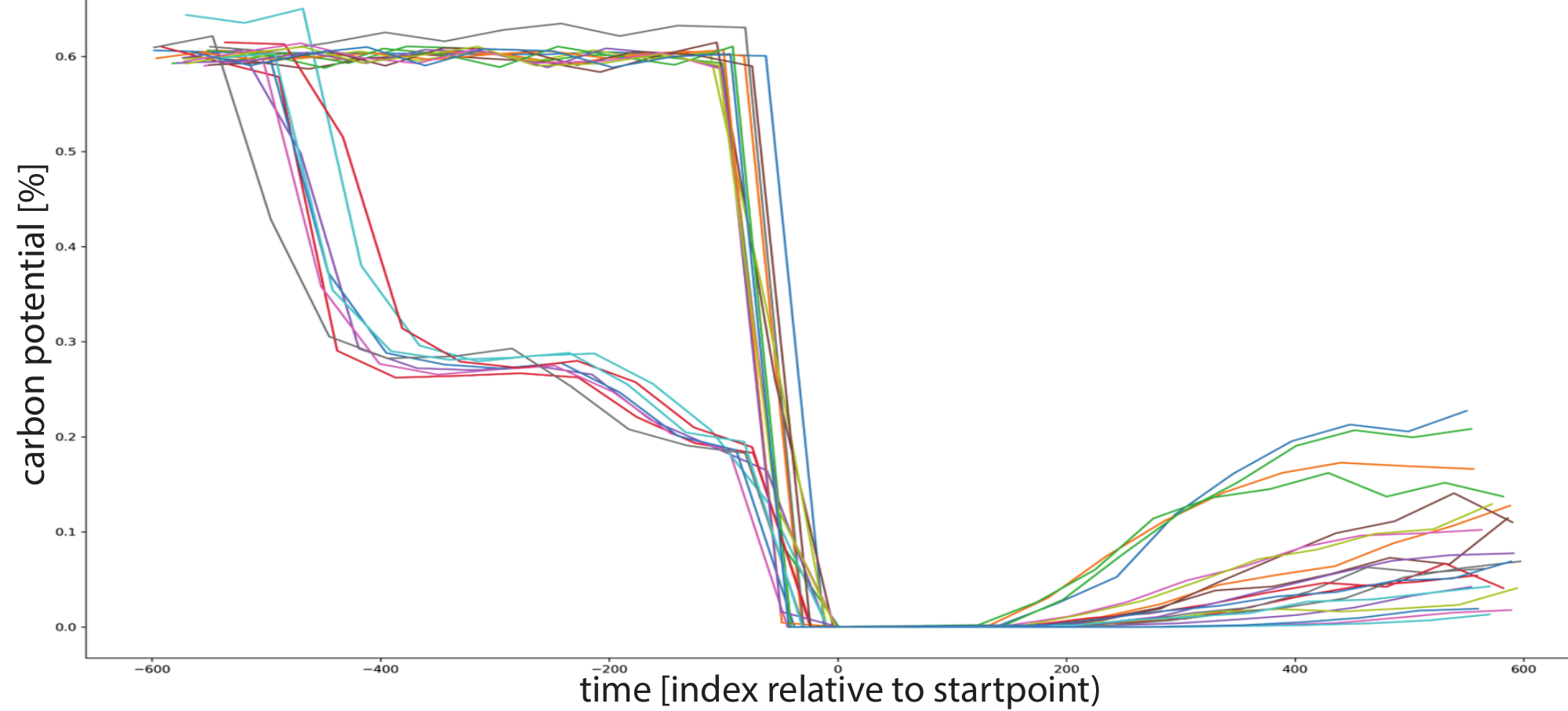}
\caption{Raw sensor values visualized at the process start}
\label{fig:rawvalues}
\end{figure}

\begin{figure}[h]
\includegraphics[width=\linewidth]{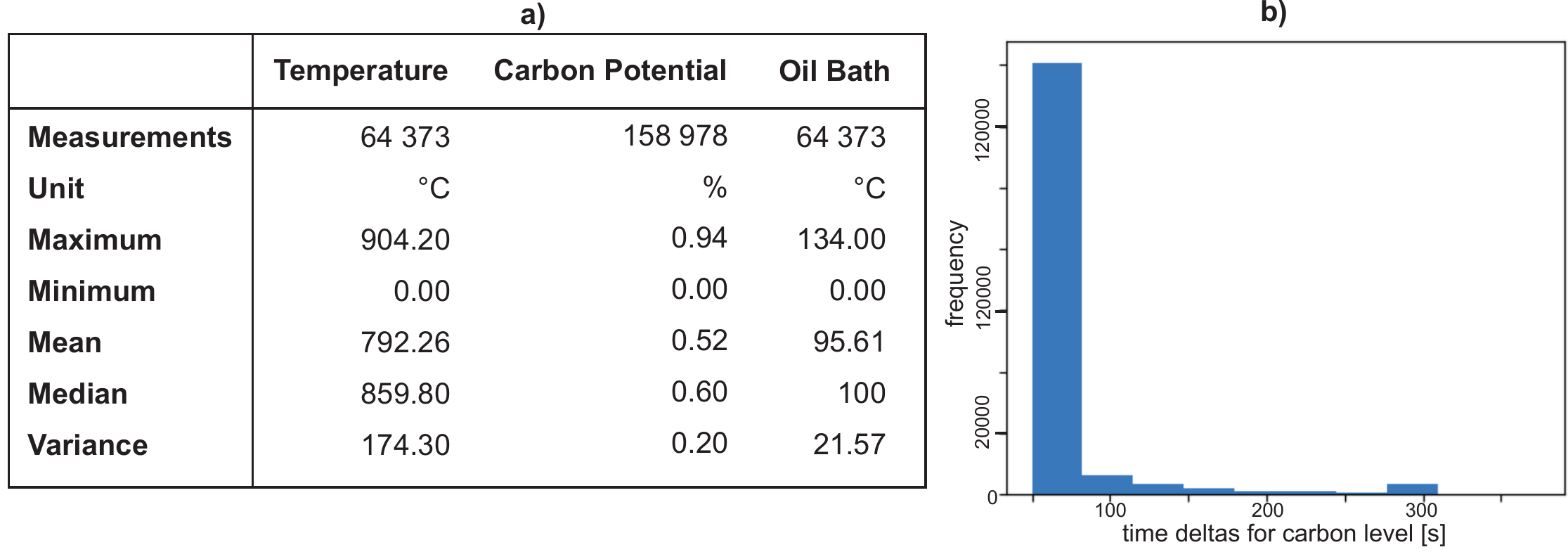}
\caption{\textbf{a)} Data set overview \textbf{b)} Histogram of sampling rate from carbon potential}
\label{fig:statistics}
\end{figure}

In comparison to complex deep architectures, 1D convolutional neural networks (1D-CNNs) provide the possibility for real-time fault detection \cite{Eren2018} and sensor analysis. The feature extraction and selection are done implicitly by the network itself. In this context, the handling of sensor readings is essential to build enhanced applications for monitoring and prediction purposes \cite{Kammerer2019}. In this work, we detect and restore production phases of a carburizing furnace using the carbon potential inside the furnace. While automated furnaces offer a built-in functionality to record start and end of a production phase, manually operated furnaces do not necessarily provide such a function. 
\begin{figure}[b]
\includegraphics[width=\linewidth]{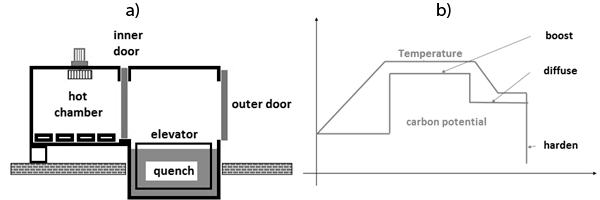}
\caption{\textbf{a)} Batch/IQ (Chamber with integrated quench) furnace \cite{Winter2015}. The load is put into the hot chamber over the quench with an elevator located at the bottom position. This approach allows for reloading the hot chamber while the previous load is still in the hardening bath (quench) \textbf{b)} Typical process setup for carburizing}
\label{fig:process}
\end{figure}
The aim of this approach is to upgrade simple machines, which only record raw sensor data, with a feature to identify the production phases based on gathered sensor stream data. Therefore, a model is trained that classifies a production phase by recognizing its start and end point. Hereby, the production workflow is accomplished as follows: A new load waiting on top of the elevator is moved into the hot chamber (see Figure \ref{fig:process}a). Note that opening the inner furnace door and pushing the cold load into the hot chamber causes changes in the gathered sensor values (i.e., a pattern can be observed, see Figure \ref{fig:rawvalues}). When the material is heated up and kept on a specific temperature and potential for a given time, then, at the end of the cycle, when it is rapidly cooled down using an oil quench, the transfer from the hot chamber to the quench will once more cause a change in the gathered sensor values (again, a pattern can be observed, see Figure \ref{fig:process}b).
The statistics for the three used sensor variables can be seen in Fig. \ref{fig:statistics}a. Note that we used three months of recorded sensor data, where the carbon value is recorded at a sample rate of mostly one value per minute (see Figure \ref{fig:statistics}b). During the recorded three months, 212 production phases were conducted. These phases are denoted as charges. Notably, the process usually takes up to four hours. 

\section{The Application}
\begin{figure}[h]
\includegraphics[width=\linewidth]{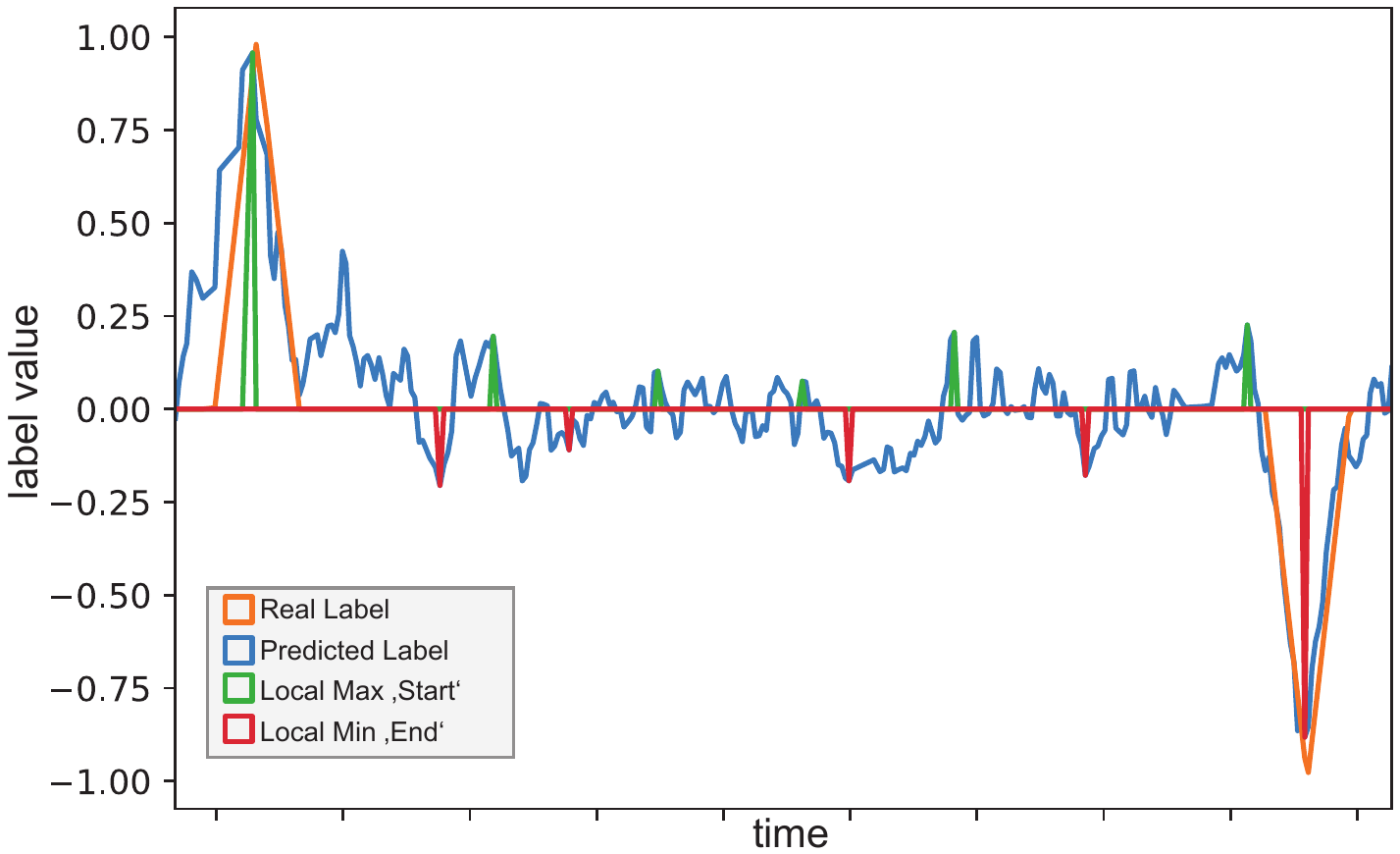}
\caption{Generated label values with prediction and filtering}
\label{fig:labels}
\end{figure}

\begin{figure}[h]
\includegraphics[width=\linewidth]{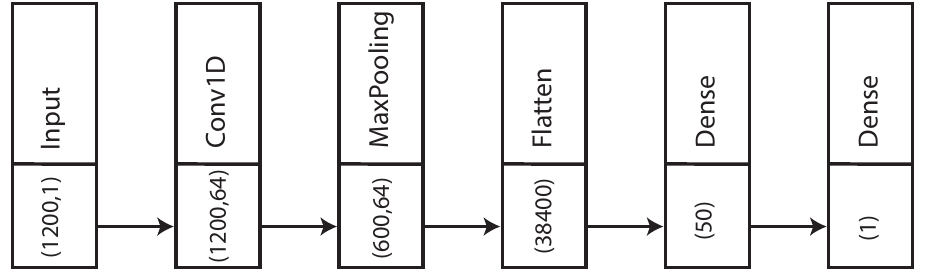}
\caption{Architecture of the 1D-CNN}
\label{fig:architecture}
\end{figure}

Before the actual machine learning can be applied, a preprocessing is required. First of all, data is stored in a SQL database. For a simplified workflow using python \cite{Brownlee2018}, the data is transferred into a CSV format. For each charge, we know the timestamps of the start and end point. Therefore, we can introduce the tree class labels (\textit{start}, \textit{end}, \textit{between}) and use them to apply a supervised approach to train a machine learning model. The task is to assign the correct classification of the current input sensor values, which can be expressed via \textit{sliding windows}. For example, a sliding window of size 1200 values, which corresponds to a record length of about 20 minutes, is shifted over the sensor data stream. To each sliding window, a label is assigned. The label for the whole sliding window is denoted as the label of the corresponding label of the centered value in the window. However, we transferred the classification problem into a regression problem to express the proximity of the current sliding window to the next phase. Hereby, the label value $l$ can be interpreted as follows.
\begin{equation}
l = 
  \begin{cases} 
   \text{start} & \text{if } l \approx 1 \\
   \text{between}       & \text{if } l \approx 0 \\
   \text{end}       & \text{if } l \approx -1
  \end{cases}
\end{equation}
Before preprocessing was applied, the labels \textit{start} and \textit{end} were assigned only at the exact position of a phase change. After processing, the values in the area of a phase change can be marked as different, as they differ from value $0$, which means a change was detected. In contrast, in the case of a \textit{start}, the generated label corresponds exactly to value $1$ at the exact timestamp $t_{ex}$, for which the charge started and is reduced linearly along the sliding window $sw$ of duration $t_{sw}$. Hereby, the new label $l_{new}$, for a sensor measurement with value $v_{sm}$ and time $t_{sm}$, is assigned as follows:
\begin{small}
\begin{equation}
distance = abs (t_{ex} - t_{sm}),
l_{new} = 1-(distance/(t_{sw}/2)
\end{equation}
\end{small}
The label workflow can be seen in Fig. \ref{fig:labels}. First, the \textit{real labels} are generated on a scale from $-1$ to $+1$. Based on these labels, a 1D-CNN is trained. The results can be seen in the curve \textit{predicted label}. To transfer the results of the predicted curve back to a unique timestamp, local maxima and minima are calculated. Finally, a threshold of $0.5$ ensures that only strong peaks are accepted. It is evident that the predicted label fits well to the real label. Therefore, the procedure seems suitable for the detection of production phases.
The machine learning model is trained using the framework Keras \cite{Keras2015}. We used the network architecture presented in Fig. \ref{fig:architecture}. Hereby, the input is provided by the current sliding window. Next, a convolution layer with 64 filters of kernel size 3 is applied and summarized in another layer using \textit{max pooling}. The generated matrix from the filter results is then flattened and serves as the input for a dense network. By using the filter-based approach, the network automatically weights more important features. To validate the approach, the data set is split up with a 80/20 ratio of train and test set. With the used network, we are able to restore the timestamps with a mean distance to the actual starting point of 93.12 seconds without any false positive detection.

\section{Summary and Outlook}
In this paper, we showed a workflow how to use 1D convolutional neural networks (1D-CNNs) to recognize production phases based on raw sensor value input. Based on simple timestamps for start and end points, this classification is then transformed into a regression problem. Finally, the results are filtered to receive the best fitting prediction. The approach showed a good precision and can be used for further analysis. In future approaches, similar procedures could be applied to detect more complex states and use these states as input for advanced analytics, such as predictive maintenance \cite{Hoppenstedt}.

\end{document}